\documentclass[conference]{IEEEtran}
\IEEEoverridecommandlockouts
\usepackage{cite}
\usepackage{amsmath,amssymb,amsfonts}
\usepackage{algorithmic}
\usepackage{graphicx}
\usepackage{textcomp}
\usepackage{xcolor}
\usepackage{booktabs}
\usepackage{multirow}
\usepackage{array}
\usepackage{times}
\usepackage{authblk}

\def\BibTeX{{\rm B\kern-.05em{\sc i\kern-.025em b}\kern-.08em
    T\kern-.1667em\lower.7ex\hbox{E}\kern-.125emX}}
\begin{document}

% \title{Exploring Linear Attention Alternative for Single Image Super-Resolution*\\
\title{Exploring Linear Attention ALternative for Single Image Super-Resolution\\
}

\author[1]{Rongchang Lu}
\author[2]{Changyu Li}
\author[2]{Donghang Li}
\author[23*]{Guojing Zhang}
\author[2**]{Jianqiang Huang}
\author[4]{Xilai Li}

\affil[1]{\textit{\small School of Ecological and Environmental Engineering, Qinghai University, Xining, 810016, China}}
\affil[2]{\textit{\small School of Computer Technology and Application, Qinghai University, Xining, 810016, China}}
\affil[3]{\textit{\small Qinghai Provincial Laboratory for Intelligent Computing and Application, Qinghai University, Xining, 810016, China}}
\affil[4]{\textit{\small College of Agriculture and Animal Husbandry, Qinghai University, Xining, 810016, China}}
\affil[*]{\small zhanggj@qhu.edu.cn, hjqxaly@163.com}

\maketitle

\begin{abstract}
Deep learning-based single-image super-resolution (SISR) technology focuses on enhancing low-resolution (LR) images into high-resolution (HR) ones. Although significant progress has been made, challenges remain in computational complexity and quality, particularly in remote sensing image processing. To address these issues, we propose our Omni-Scale RWKV Super-Resolution (OmniRWKVSR) model which presents a novel approach that combines the Receptance Weighted Key Value (RWKV) architecture with feature extraction techniques such as Visual RWKV Spatial Mixing (VRSM) and Visual RWKV Channel Mixing (VRCM), aiming to overcome the limitations of existing methods and achieve superior SISR performance. Our work demonstrates the ability to provide effective solutions for high-quality image reconstruction. Under the 4x Super-Resolution tasks, compared to the MambaIR model, we achieved an average improvement of 0.26\% in PSNR and 0.16\% in SSIM.
\end{abstract}


\section{Introduction}
Single Image Super-Resolution (SISR) technology holds a significant position in the field of image processing, aiming to reconstruct LR images into HR images to meet the increasing demand for image clarity. SISR has broad application potential in various fields such as medical imaging\cite{medical}, remote sensing\cite{remotesensing}, and video surveillance\cite{video}. HR images not only provide more detailed visual information but also significantly enhance the accuracy of decision support, such as in medical diagnosis and environmental monitoring applications \cite{sisr1,sisr2}.

Recent deep learning-based SISR methods have made significant strides in reconstructing LR images by training models on large numbers of LR/HR image pairs. Convolutional Neural Networks (CNNs) are commonly used for feature extraction, enabling detailed image recovery through multiple nonlinear transformations \cite{image-restoration1,image-restoration2}. However, challenges remain in applying SISR to remote sensing images, such as localized features, complex structures, scale differences, and noise \cite{computervision}. To address these, several advanced models have been proposed. The SwinIR model leverages hierarchical feature representation and a shift-window-based self-attention mechanism for long-range dependency capture \cite{swinir}. SwinFIR further enhances this by integrating a Fast Fourier Convolution (FFC) component for improved global information efficiency \cite{swinfir}. The MambaIR model, which applies the Mamba model for image restoration, combines convolution with channel attention mechanisms to improve representation \cite{mambair,vision-mamba}. The Hybrid Attention Transformer (HAT) optimizes resource allocation and enhances feature extraction by selectively applying attention \cite{hat}. The RWKV model\cite{rwkv}, particularly its sixth generation (RWKV v6), improves training stability and convergence speed by integrating Recurrent Neural Networks (RNNs) and Transformer strengths while avoiding full attention mechanisms with linear complexity\cite{eagleandfinch}.

This study proposes a novel approach that combines the RWKV architecture with advanced feature extraction techniques and verified through ablation experiments, aiming to overcome the limitations of existing attention-based methods, such as the quadratic complexity. The main contributions of this study are as follows.

\subsection{Methodological Innovations}
\subsubsection{Feed Forward Network}

ChannelMix is superior to traditional MLP in improving channel-wise feature mixing and improving information flow, thus outperforming MLP in feature extraction and image reconstruction quality.

\subsubsection{Omni-Quad Shift}

To capture long-range dependencies, we introduce a complementary mechanism: Omni-Quad Shift. It enables models to effectively capture multi-scale features and spatial transformations.

\subsubsection{Weighted Key Value (WKV) 2D Scanning}

Conventional linear models can only capture the one-dimensional information, which is also quoted "scanning". Despite the 2D scanning of state-space model, the quality and performance still demands improvement. To address this issue, we propose the 2D scanning version of the WKV mechanism in this paper.

\subsection{Experimental Advances}
\subsubsection{Faster Training With SOTA Performance}

Our model reaches the highest score on popular datasets, including Set14, BSD100, and Urban100 with approximately 15\% less training time compared to MambaIR.

\section{Related Work}

\subsection{Attention-based Super Resolution}
\begin{figure*}[!t]
    \centering
    \includegraphics[width=\textwidth, keepaspectratio]{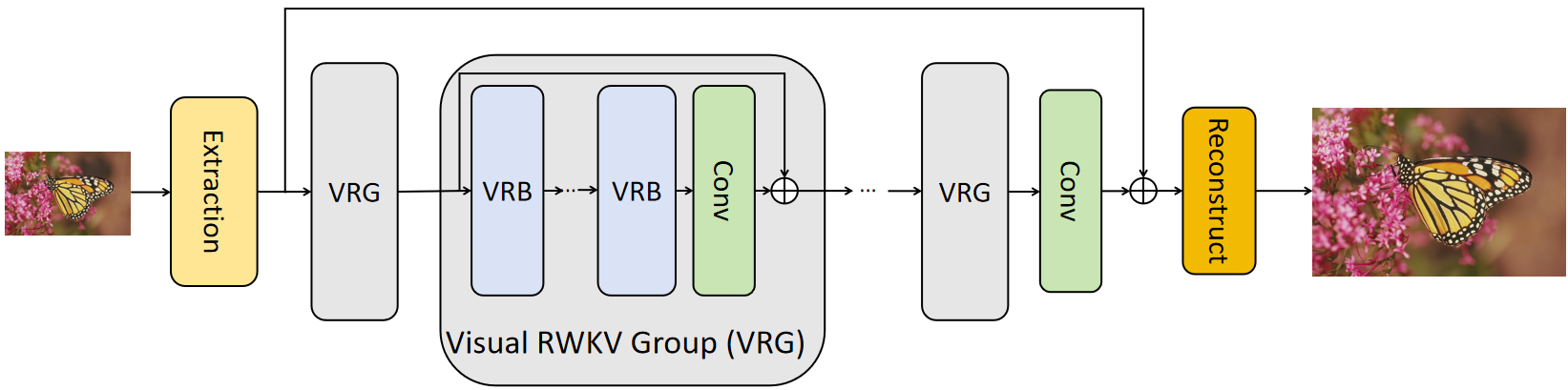}
    \caption{The architecture of OmniRWKVSR.}
    \label{arch}
\end{figure*}

In recent years, attention mechanisms have significantly advanced SISR by enabling models to effectively capture long-range dependencies and focus on critical image regions. Notably, the Swin Transformer-based model, SwinIR \cite{swinir}, has demonstrated exceptional performance across various image restoration tasks, including super-resolution, denoising, and JPEG artifact reduction \cite{denoising, JPEGartifactreduction}.

% SwinIR \cite{denoising, JPEGartifactreduction}leverages hierarchical feature representation and shifted window-based self-attention to model long-range dependencies while maintaining computational efficiency. Further extending the capabilities of SwinIR, the SwinFIR \cite{swinfir} model incorporates FFC components to achieve an image-wide receptive field. This enhancement allows SwinFIR to capture global context more effectively, leading to improved performance in image super-resolution tasks. 

Despite the significant advancements brought by attention mechanisms in SISR \cite{visiontransformer, swinir}, a notable limitation persists: the computational complexity of self-attention operations scales quadratically with the input size. This quadratic complexity arises because the attention mechanism computes pairwise interactions between all elements in the input sequence, leading to substantial computational and memory demands, especially for HR images. 

\subsection{State Space Model (SSM)}

To address the computational challenges of attention-based SISR, several models\cite{swinir, swinfir} have integrated innovative mechanisms to improve efficiency. The MambaIR model\cite{mambair} introduces a Selective State Space 2D (SS2D) mechanism\cite{s4, mamba}, which employs SSM with selective scanning strategies to capture long-range dependencies while maintaining linear computational complexity relative to the input size. This design effectively reduces the computational burden of traditional quadratic attention mechanisms, making MambaIR scalable and efficient for HR image restoration tasks. The SS2D mechanism allows MambaIR to model intricate image details without incurring the prohibitive costs typical of self-attention methods, thus balancing performance and efficiency for large-scale image processing.

% On the other hand, the HAT model \cite{hat}, building on the RWKV architecture, combines the strengths of both attention and non-attention mechanisms. HAT selectively applies attention to optimize resource allocation, reducing computational overhead while enhancing feature extraction. Through meta-learning, HAT adapts its attention mechanism dynamically, improving generalization and speeding up training. This enables HAT to effectively capture both local and global features, excelling in tasks that require understanding fine-grained details while maintaining computational efficiency.

\subsection{RWKV Structure}

Despite the advancements introduced by MambaIR \cite{mambair} and the SS2D \cite{vision-mamba} mechanism, challenges remain in terms of prolonged training times, unstable performance metrics, rapid convergence, and suboptimal feature extraction capabilities. The RWKV v6\cite{rwkv} offers solutions to these issues by integrating the strengths of RNNS and Transformers while eliminating the reliance on attention mechanisms. One of the key improvements in RWKV v6 is its improved training stability and convergence speed. By employing careful initialization strategies and optimized training protocols, RWKV v6 mitigates issues related to training instability and premature convergence, leading to more robust and reliable performance across various tasks. Additionally, RWKV's linear computational complexity with respect to input sequence length addresses the inefficiencies associated with quadratic complexity in traditional attention mechanisms. Hence, in this paper, we propose this innovative approach that integrates the strengths of the RWKV architecture with advanced feature extraction techniques to enhance SISR tasks.

\section{Method}

This section first details the overall pipeline of our proposed OmniRWKVSR model and then explains the basic block of the network: Visual RWKV Residual Group (VRG).

\subsection{OmniRWKVSR Model}\label{OM}

The overall structure of Omni-Scale RWKV model is shown in Fig. \ref{arch}. We denote the LR input image as $\boldsymbol{I_{LR}} \in \mathbb{R}^{B \times C \times H \times W}$, and the HR target image as $\boldsymbol{I_{HR}} \in \mathbb{R}^{B \times C \times H \times W}$. 

\begin{figure*}[!t]
    \centering
    \includegraphics[width=\textwidth, keepaspectratio]{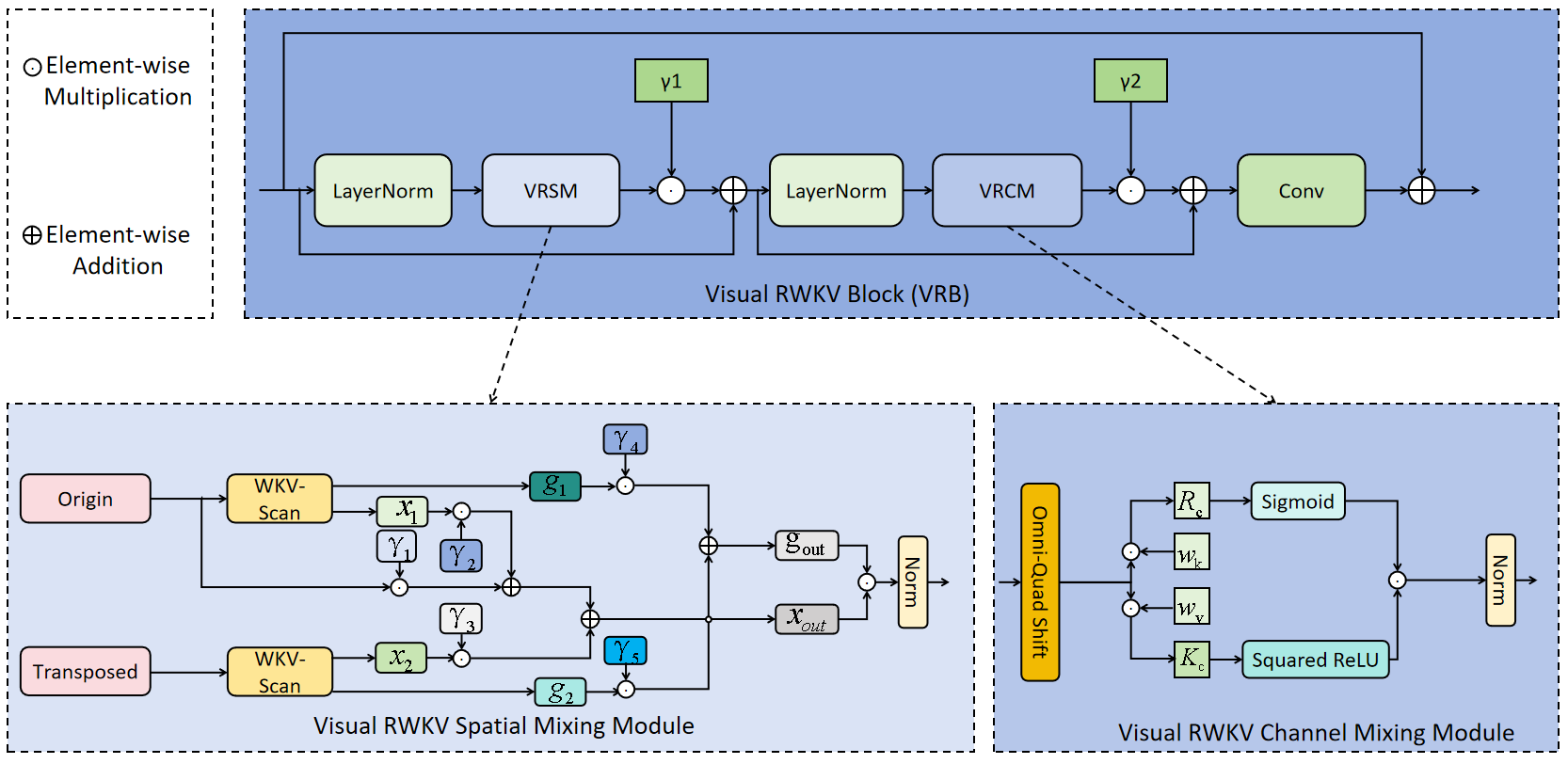}
    \caption{The architecture of VRB, VRSM and VRCM.}
    \label{arch2}
\end{figure*}

Initially, we use a $3 \times 3$ CNN layer, worked as the shallow feature extractor (Extraction), to extract low-level features from the LR input image.

Secondly, we use multiple VRGs to extract high-level features from $\boldsymbol{F}_{\text{embd}}^{1}$, denoted as $\boldsymbol{F}_{\text{feat}}$:

\begin{equation}
\boldsymbol{F}_{\text{embd}}^{n} = f_{VRG}^{n}(f_{VRG}^{n-1}( \ldots f_{VRG}^{1}( \boldsymbol{F}_{\text{embd}}^{1}))),
\end{equation}
where $f_{VRG}^{n}$ is a VRG, $\boldsymbol{F}_{\text{embd}}^{0}$ is the output of the shallow feature extractor, and $f_{VRG}^{n-1}( \ldots f_{VRG}^{1}( \boldsymbol{F}_{\text{embd}}^{0}))$ is the output of the previous VRG.

Thirdly, we use a $3 \times 3$ CNN layer and a one-step upsampling operation (Reconstruct) to upsample $\boldsymbol{F}_{\text{embd}}$ to the HR resolution, denoted as $\boldsymbol{F}_{HR}$:

\begin{equation}
\boldsymbol{F}_{HR} = \text{upsample}(\text{conv}_{\text{feat}}(\boldsymbol{F}_{\text{embd}}^n)),
\end{equation}
where $\text{conv}_{\text{feat}}$ is a $3 \times 3$ CNN layer, $\boldsymbol{F}_{\text{embd}}^n$ is the output of the n-th VRG, and $\text{upsample}$ is a one-step upsampling operation.
% \begin{figure*}[!t]
%     \centering
%     \includegraphics[width=\textwidth, keepaspectratio]{Fig1.png}
%     \caption{The architecture of our work.}
% \end{figure*}

% \begin{figure*}[!t]
%     \centering
%     \includegraphics[width=\textwidth, keepaspectratio]{Fig2.png}
%     \caption{The architecture of VRG.}
% \end{figure*}

Finally, we add the mean and variance back to $\boldsymbol{F}_{HR}$ and obtain the high-quality reconstructed image, denoted as $\boldsymbol{I}_{HR}$:

\begin{equation}
\boldsymbol{I}_{HR} = \mu_{LR} + \sigma_{LR} \odot \boldsymbol{F}_{HR},
\end{equation}
where $\odot$ is the element-wise multiplication operator, $\mu_{LR}$ and $\sigma_{LR}$ are the mean and variance of the LR input image, respectively.

\subsection{Visual RWKV Group}

A VRG is a collection of residual blocks that integrates the VRSM and VRCM as its core components. The VRSM facilitates the blending of feature maps across different resolutions, while the VRCM confines these features to specific channels. VRCM replaces traditional MLP by computing cross-channel weights via depthwise convolution. A VRG is composed of multiple Visual RWKV Residual Blocks (VRBs), and the overall structure of the VRB is shown in Fig. \ref{arch2}.

The VRG receives input from either the output of the previous VRG or the shallow feature extractor, and its output is directed to the subsequent VRG or the process for high-quality reconstruction. Features entering VRSM are initially subjected to a shift operation using Omni-Quad Shift, yielding a set of shifted feature maps that encapsulate the features of neighboring pixels. These maps are then integrated through a channel-wise linear attention mechanism within the Spatial Mixing operation, which selects significant channels and conducts 2D scan operations to reassmble the feature maps. The outcome is fed back into the input feature map through a residual connection. Following this, the output is directed to VRCM, which confines the feature maps to specific channels, with its output also reintegrated via a residual connection. The VRCM output is then passed through a $3 \times 3$ convolutional layer, added back to the input feature map through a residual connection, and advanced to the next residual block or the high-quality reconstruction phase.

% \begin{figure}[!t]
%     \centering
%     \includegraphics[width=0.5\textwidth]{Fig4.png}
%     \caption{The architecture of Omni-Quad Shift}
% \end{figure}

% \begin{figure*}[!t]
%     \centering
%     \includegraphics[width=\textwidth, keepaspectratio]{Fig1.png}
%     \caption{The architecture of OmniRWKVSR.}
% \end{figure*}

% \begin{figure*}[!t]
%     \centering
%     \includegraphics[width=\textwidth, keepaspectratio]{Fig2.png}
%     \caption{The architecture of VRB, VRSM and VRCM.}
% \end{figure*}

\begin{figure}[h]
    \centering
    \includegraphics[width=0.4\textwidth, keepaspectratio]{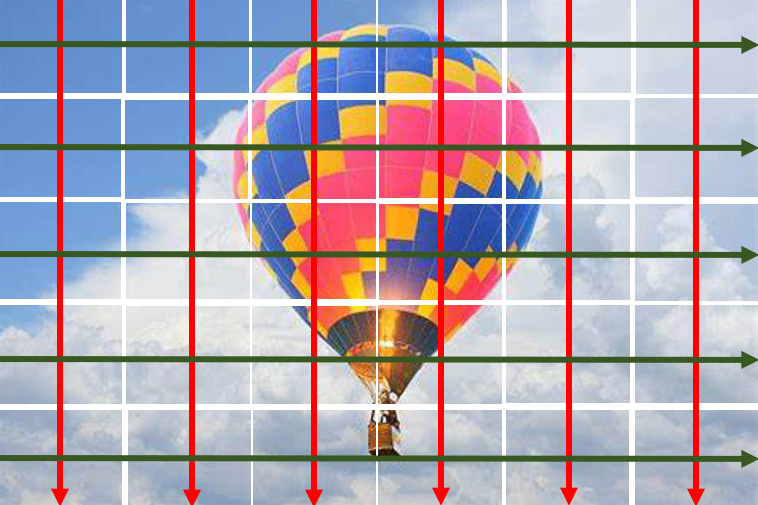}
    \caption{Visual illustration for WKV2DScan.}
    \label{visual}
\end{figure}

\subsection{WKV2DScan}

The WKV2DScan module is a two-stage process that operates on an input tensor to yield feature maps and attention weights.  The visual illustration of WKV2DScan is showed as Fig. \ref{visual}, where the scanning operation is performed horizontally and vertically to obtain the features in both directions. In each stage of WKV2DScan, a WKV-Scan operation is performed. WKV-Scan initiates with a shift operation to gather contextual data, defining the shifted tensor as $\boldsymbol{x}_{\text{ssh}}$:

\begin{equation}
\boldsymbol{x}_{\text{ssh}}^{n} = f_{\text{Omni-Quad Shift}}(\boldsymbol{x}_{\text{in}}^{n}),
\end{equation}
which is followed by a weighted sum calculation, combining the shifted tensor with the original to produce $\boldsymbol{x}_{\text{ss}}$.

\begin{equation}
\boldsymbol{x}_{\text{ss}}^{n} = \boldsymbol{x}_{\text{ssh}}^{n} \odot \text{tma}_{x} + \text{tma}_{xb},
\end{equation}
where $\boldsymbol{tma}_{\text{x}}$ is the time-dependent moving average of the input tensor, and $\boldsymbol{tma}_{\text{xb}}$ is the bias term.

Subsequently, the module calculates key parameters for the scan operation. These include learnable weights and time-dependent moving averages, which are determined by the formulas:

\begin{equation}
\begin{pmatrix}
\boldsymbol{r}_{s}^{n} \\
\boldsymbol{k}_{s}^{n} \\
\boldsymbol{v}_{s}^{n} \\
\boldsymbol{g}_{s}^{n} \\
\boldsymbol{ww}_{s}^{n}
\end{pmatrix}
=
\begin{pmatrix}
\boldsymbol{w}_{r} \\
\boldsymbol{w}_{k} \\
\boldsymbol{w}_{v} \\
\boldsymbol{w}_{g} \\
\boldsymbol{w}_{\text{ww}}
\end{pmatrix}
% \cdot
\left(
\boldsymbol{x}_{\text{in}}^{n}
+
\boldsymbol{x}_{\text{ssh}}^{n}
% \odot
\left(
\begin{pmatrix}
\text{tma}_{r} \\
\text{tma}_{k} \\
\text{tma}_{v} \\
\text{tma}_{g} \\
\text{tma}_{r}
\end{pmatrix}
\oplus
\boldsymbol{x}_{\text{ss}}^{n}
\right)
\right),
\end{equation}
where $\boldsymbol{w}_{\text{r}}$, $\boldsymbol{w}_{\text{k}}$, $\boldsymbol{w}_{\text{v}}$, $\boldsymbol{w}_{\text{g}}$, $\boldsymbol{w}_{\text{ww}}$are the learnable weights that is shared during two scans, while the $\boldsymbol{tma}_{\text{r}}$, $\boldsymbol{tma}_{\text{k}}$, $\boldsymbol{tma}_{\text{v}}$, $\boldsymbol{tma}_{\text{g}}$, are ones that exist on their respective scans, so that not only are they dependent, but also can be learned from either scan during training.

The scanned feature map is then derived using the WKV6 module, with the formula:

\begin{equation}
\boldsymbol{x}_{\text{out}}^{n} = \text{WKV6}(H, \boldsymbol{r}, \boldsymbol{k}, \boldsymbol{v}, \boldsymbol{w}, \boldsymbol{u}),
\end{equation}
where $H$ represents the head size.

The first WKV-Scan operation is applied to the input feature map $\boldsymbol{F}_{\text{embd}}^{n}$, yielding the first set of scanned feature maps and attention weights $\boldsymbol{x}_{\text{o1}}^{n}$, $\boldsymbol{g}_{\text{o1}}^{n}$. For the second operation, the transposed input feature map ${\boldsymbol{F}_{\text{embd}}^n}^T$ is used to obtain the second set $\boldsymbol{x}_{\text{o2}}^{n}$, $\boldsymbol{g}_{\text{o2}}^{n}$.

Finally, the WKV2DScan module outputs a tuple of the scanned feature maps and attention weights, computed as:

\begin{equation}
\boldsymbol{x}_{\text{out}}^{n} = \boldsymbol{F}_{\text{embd}}^{n} \odot \gamma_{s1}^{n} + \boldsymbol{x}_{o1}^{n} \odot \gamma_{s2}^{n} + \boldsymbol{x}_{o2}^{n} \odot \gamma_{s3}^{n},
\end{equation}

\begin{equation}
\boldsymbol{g}_{\text{out}}^{n} = \text{Sigmoid}(\boldsymbol{g}_{o1}^{n} \odot \gamma_{s4}^{n} + \boldsymbol{g}_{o2}^{n} \odot \gamma_{s5}^{n}),
\end{equation}

\begin{equation}
\boldsymbol{y}_{\text{out}} = \boldsymbol{x}_{\text{out}}^{n} \odot \boldsymbol{g}_{\text{out}}^{n},
\end{equation}
where $\gamma_{s1}^{n}, \gamma_{s2}^{n}, \gamma_{s3}^{n}, \gamma_{s4}^{n}, \gamma_{s5}^{n}$ are learnable weights, and $\boldsymbol{y}_{\text{out}}$ is the output of VRSM.

\begin{figure}[h]
    \centering
    \includegraphics[width=0.4\textwidth]{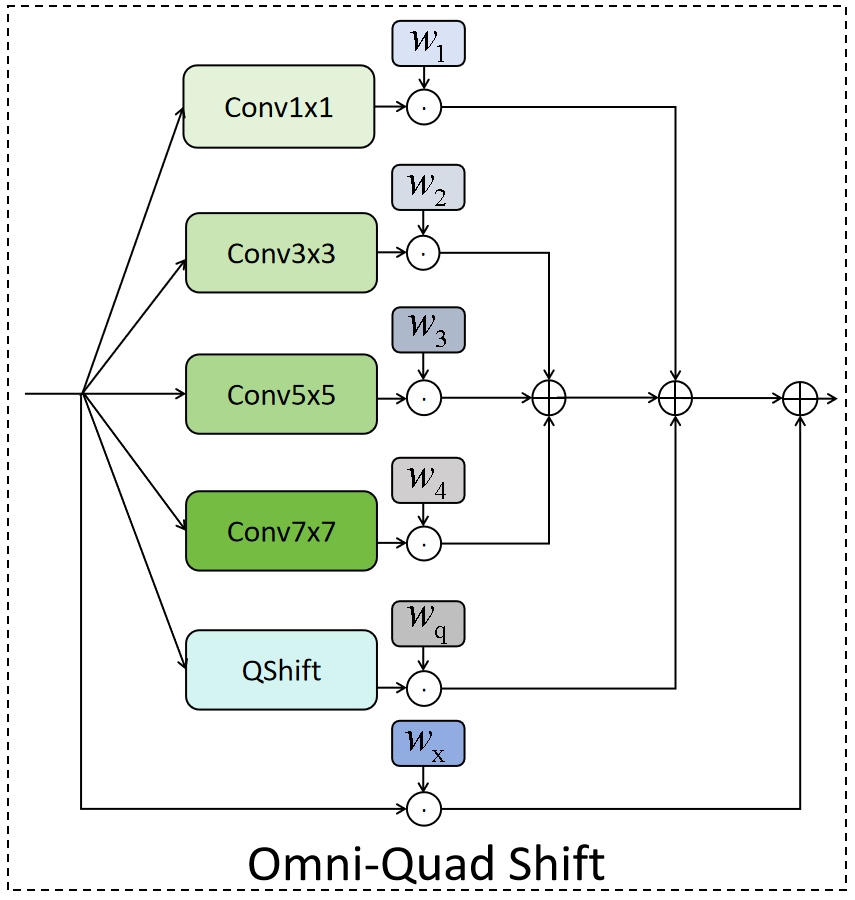}
    \caption{The architecture of Omni-Quad Shift.}
    \label{omni}
\end{figure}

\subsection{Omni-Quad Shift}

Our Omni-Quad Shift leverages multi-scale convolutional kernels ($1\times 1$, $3\times 3$, $5\times 5$, and $7\times 7$) to aggregate information at different receptive field sizes. Each convolutional kernel is depthwise and dilated, capturing both local and global features. The outputs are combined using learnable weights to produce a unified representation. Of these convolutions, $\text{Conv}_{1 \times 1}(\boldsymbol{X})$ captures point-wise features, $\text{Conv}_{3 \times 3}(\boldsymbol{X})$ extracts local spatial details, $\text{Conv}_{5 \times 5}(\boldsymbol{X})$ enhances mid-range dependencies and $\text{Conv}_{7 \times 7}(\boldsymbol{X})$ expands the receptive field for global context. Also, learnable weights $w_x, w_1, w_2, w_3, w_4, w_q$ modulate the contribution of each convolution and Quad-directional Shift (QShift):

\begin{equation}
\boldsymbol{Y} = \begin{pmatrix}
w_x \\
w_1 \\
w_2 \\
w_3 \\
w_4 \\
w_q
\end{pmatrix}^T
\begin{pmatrix}
\boldsymbol{X} \\
\text{Conv}_{1 \times 1}(\boldsymbol{X}) \\
\text{Conv}_{3 \times 3}(\boldsymbol{X}) \\
\text{Conv}_{5 \times 5}(\boldsymbol{X}) \\
\text{Conv}_{7 \times 7}(\boldsymbol{X}) \\
\text{QShift}(\boldsymbol{X})
\end{pmatrix},
\end{equation}

After obtaining the output tensor $\boldsymbol{Y}$, we rearrange the output back into the original spatial format.

\section{Experiment}

\subsection{Dataset and implementation details}

The training and validation set consisted of 2650 images from Flickr2K\cite{flckr2k}, 800 images from DIV2K\cite{div2k}, and 2800 images from RSSCN7\cite{rsscn7} total, of which the training set occupies 95\%. In the Flickr2K and DIV2K dataset (DF2K), the image resolution is 2048 × 2048 and 2048 × 1080 pixels. In the RSSCN7 dataset, the image resolution is 400 × 400 pixels. The resolution of LR in the training set is fixed to 48 × 48, while that of HR is 96 × 96 at a 2x scale factor and 192 × 192 at 4x one. Our experiments were carried out to ensure that all trained hyperparameters and the number of parameters were approximately the same. Our models were evaluated on widely used benchmark datasets: Set5\cite{set5}, Set14\cite{set14}, BSD100\cite{bsd100}, Urban100\cite{urban100} and Manga109\cite{manga109}. We also performed tests on our own Gangcha MouseHole (GMH) remote sensing dataset, the validation set of the DF2K dataset, and the public RSSCN7 remote sensing dataset.

The GMH dataset is captured by a DJI Mavic 3E drone in Qinghai Province, China, which includes 325 images of mouse holes in the middle reaches at an altitude of 30 meters, 222 images of mouse holes in the upper reaches at an altitude of 40 meters, and 60 images of mouse holes in the middle and upper reaches at an altitude of 80 meters over the grasslands of Gangcha County. In total, there are 607 images, each with a resolution of 5280 × 3956 pixels split into 72 × 72 LR pieces to perform SR tasks.

RSSCN7\cite{rsscn7} is a remote sensing image dataset for scene classification, featuring 2800 high-resolution RGB images at 400 × 400 pixels. It includes seven classes: grassland, forest, farmland, industrial/commercial, river/lake, residential, and parking lots, with 400 images each.

\begin{figure*}[!t]
    \centering
    \includegraphics[width=0.9\textwidth, keepaspectratio]{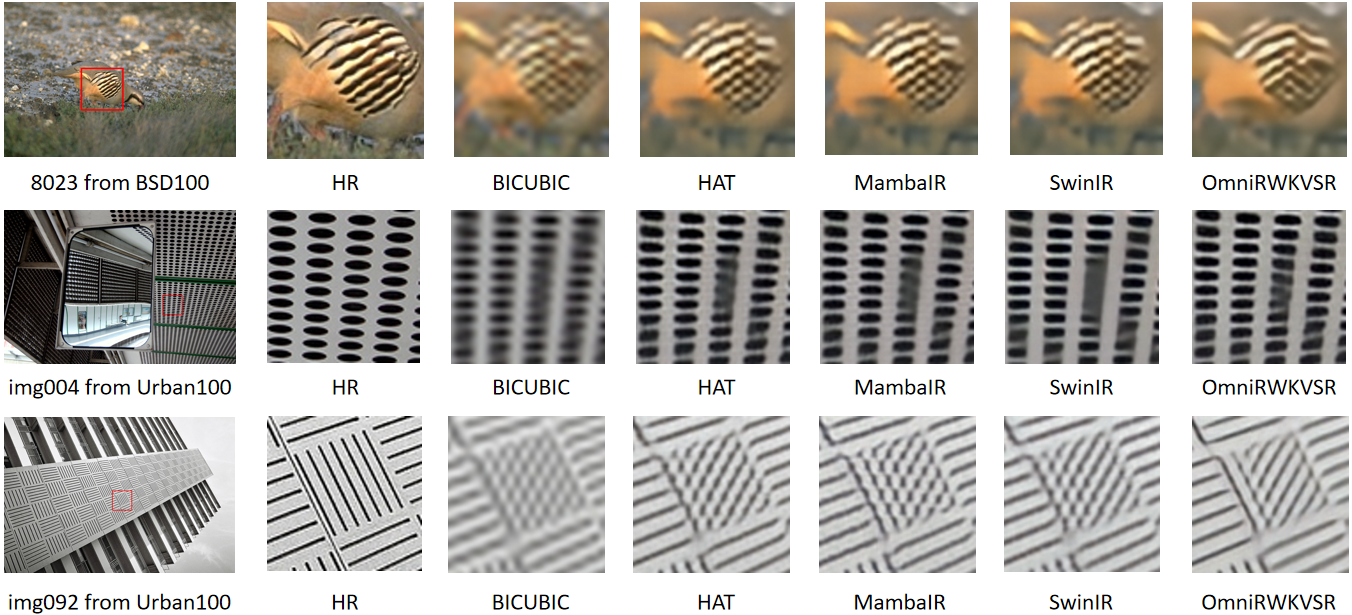}
    \caption{Visual results on benchmark datasets for $\times 4$ super-resolution.}
    \label{result}
\end{figure*}

\begin{figure*}[!t]
    \centering
    \includegraphics[width=0.9\textwidth, keepaspectratio]{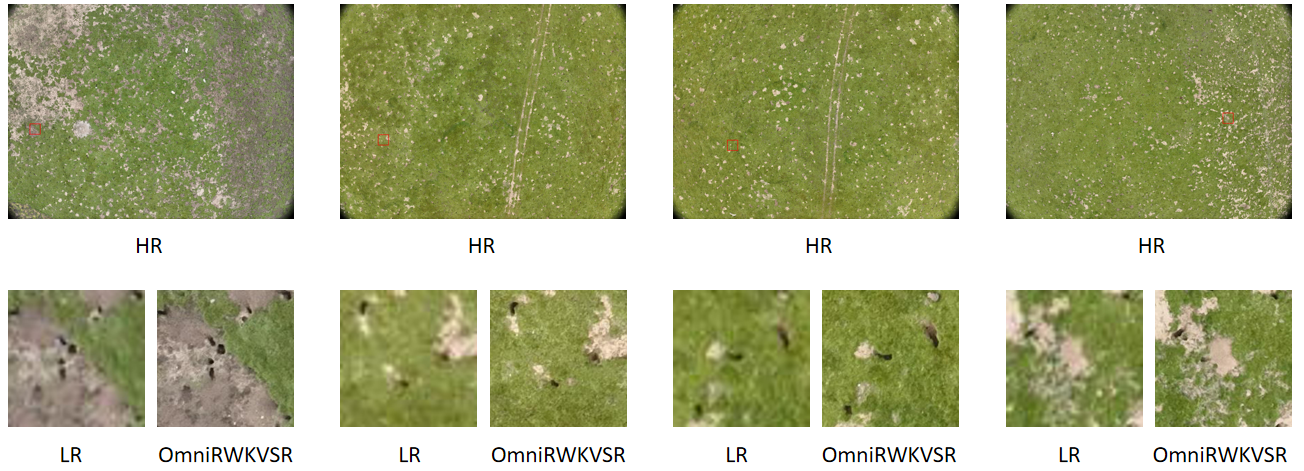}
    \caption{Visual results on Gangcha MouseHole dataset.}
    \label{GMH}
\end{figure*}

Our model is trained on RGB channels and augmented data with random rotations, sizes and flipping. The PSNR\cite{psnr} and SSIM\cite{ssim} on the Y channel in the YCbCr space is calculated as quantitative measurements. Our implementation of OmniRWKVSR contains 96 channels and 16 VRG blocks. The minibatch size is set to 16, using Adam optimizer for optimization. In the first stage, a pre-training of 20000 iterations is performed using \(\mathcal{L}\textit{1}\) Loss function, and the learning rate is set to 1e-4. We use none of learning rate scheduler so as to ensure our model can keep its performance without any tricks for training.

\subsection{Qualitative evaluations}

Fig. \ref{result} presents a qualitative comparison of the proposed method at a 4x scale factor on the Set14 and Urban100 datasets. The figure displays the visual effects of HR, BICUBIC, HAT\cite{hat}, MambaIR\cite{mambair}, SwinIR\cite{swinir}, and OmniRWKVSR (our work) on the specified images.

Taking “img092” as an example, both MambaIR and SwinIR have issues in restoring stripe textures, where the stripes appear blurry and interlaced. Our method also faces challenges in achieving high-quality super-resolution reconstruction due to the significant difference between the test set and the training set. However, OmniRWKVSR performs relatively better in this regard. It can more accurately recover the characteristics of stripes, making them clearer and more realistic compared to other methods. While ensuring lightweight complexity, OmniRWKVSR is able to effectively achieve better super-resolution reconstruction results under the given circumstances.

\begin{table*}[h!]
    \centering
    \renewcommand{\arraystretch}{1.2}
    \caption{\centering \textbf{Comparison of OmniRWKVSR and SOTA Super-Resolution Models Across Different Datasets. Bold data indicate the best method.}}
    \resizebox{1\textwidth}{!}{
    \begin{tabular}{c c c c c c c c c}
        \toprule
        Scale & Models & DF2K+RSSCN7 & Set5 & Set14 & BSD100 & Urban100 & Manga109 & Gangcha MouseHole \\ 
        & & PSNR/SSIM$\uparrow$ & PSNR/SSIM$\uparrow$ & PSNR/SSIM$\uparrow$ & PSNR/SSIM$\uparrow$ & PSNR/SSIM$\uparrow$ & PSNR/SSIM$\uparrow$ & PSNR/SSIM$\uparrow$ \\ 
        \midrule
        \multirow{7}{*}{x2} 
        & \textbf{OmniRWKVSR(ours)} & \textbf{31.86/0.8558} & \textbf{35.36}/0.9318 & \textbf{31.74}/0.8737 & 30.63/\textbf{0.8517} & \textbf{29.43/0.8860} & \textbf{35.42/0.9513} & \textbf{33.02/0.8281} \\   
        & MambaIR\cite{mambair} & 31.86/0.8555 & 35.34/0.9290 & 31.75/0.8737 & 30.60/0.8498 & 29.29/0.8801 & 35.32/0.9506 & 33.01/0.8279 \\  
        & SwinIR\cite{swinir} & 31.79/0.8536 & 35.34/0.9317 & 31.75/0.8749 & 30.60/0.8496 & 29.29/0.8775 & 35.32/0.9512 & 32.99/0.8264 \\ 
        & SwinFIR\cite{swinfir} & 31.78/0.8539 & 35.35/0.9318 & 31.76/0.8752 & 30.62/0.8502 & 29.31/0.8780 & 35.30/0.9509 & 32.98/0.8259 \\  
        & HAT\cite{hat} & 31.77/0.8547 & 35.36/\textbf{0.9319} & 31.77/0.8755 & 30.63/0.8516 & 29.38/0.8797 & 35.32/0.9514 & 33.01/0.8280 \\ 
        & MambaIRv2\cite{mambairv2} & 31.86/0.8554 & 35.28/0.9310 & 31.69/\textbf{0.8765} & \textbf{30.64}/0.8517 & 29.42/0.8837 & 35.42/0.9522 & 33.00/0.8281 \\ 
        \midrule
        \multirow{7}{*}{x4} 
        & \textbf{OmniRWKVSR(ours)} & \textbf{28.07/0.7192} & \textbf{30.14/0.8458} & \textbf{27.34/0.7264} & \textbf{26.66/0.6809} & \textbf{24.58/0.7160} & \textbf{27.89/0.8533} & \textbf{29.17/0.6358} \\
        & MambaIR\cite{mambair} & 28.04/0.7188 & 30.01/0.8445 & 27.29/0.7255 & 26.63/0.6801 & 24.50/0.7141 & 27.73/0.8501 & 29.15/0.6355 \\ 
        & SwinIR\cite{swinir} & 27.98/0.7168 & 29.99/0.8439 & 27.23/0.7257 & 26.60/0.6804 & 24.39/0.7091 & 27.43/0.8435 & 29.14/0.6354 \\ 
        & SwinFIR\cite{swinfir} & 27.98/0.7170 & 29.99/0.8442 & 27.23/0.7260 & 26.60/0.6803 & 24.41/0.7099 & 27.46/0.8442 & 29.13/0.6355 \\  
        & HAT\cite{hat} & 28.01/0.7179 & 30.08/0.8459 & 27.29/0.7276 & 26.63/0.6814 & 24.48/0.7131 & 27.61/0.8467 & 29.15/0.6357 \\  
        & MambaIRv2\cite{mambairv2} & 28.04/0.7192 & 30.11/0.8468 & 27.32/0.7285 & 26.67/0.6827 & 24.57/0.7179 & 27.80/0.8517 & 29.16/0.6358 \\ 
        \bottomrule
    \end{tabular}
    }
\end{table*}

\begin{table}[h]
    \centering
    \caption{Training and Inference Time and Operations Comparison of Different Models}
    \renewcommand{\arraystretch}{1}
    \resizebox{0.5\textwidth}{!}{
    \begin{tabular}{l c c c c}
        \toprule
        \textbf{Model} & \textbf{Training} & \textbf{Inference} & \textbf{Parameters[M]} & \textbf{FLOPs[G]} \\
        \midrule
        \textbf{OmniRWKVSR (ours)} & 3h 47min 58sec  & 19min 34sec & 2.96 & 112.46 \\
        MambaIR\cite{mambair}         & 4h 32min 12sec  & 21min 12sec & 2.87 & 121.28 \\
        MambaIRv2\cite{mambairv2}       & 5h 9min 9sec    & 22min 55sec & 3.12 & 121.26 \\
        SwinIR\cite{swinir}          & 5h 18min 45sec  & 25min 23sec & 2.78 & 154.95 \\
        \bottomrule
    \end{tabular}
    }
\end{table}

\subsection{Quantitative evaluations}

% To evaluate the effectiveness of RWKVIR in terms of the quality and efficiency of image super-resolution reconstruction, we compared it with several common super-resolution models at a scale factor of ×4, including SwinIR, SwinFIR, and MambaIR,  RWKVIR integrates local information processing convolutions with channel attention mechanisms to enhance its representation capabilities. The evaluation results presented in TABLE I indicate that both RWKVIR and MambaIRv2\cite{mambairv2} perform exceptionally well in the PSNR and SSIM metrics, particularly in the DF2K + RSSCN7 (validation set) and Set5 datasets. Furthermore, the RWKVIR model also shows strong performance in the Urban100 and Manga109 datasets. In general, both models excel in image quality and structural preservation, and future optimizations could enhance their adaptability and performance in diverse scenarios.

To comprehensively evaluate the effectiveness of OmniRWKVSR in terms of image super-resolution reconstruction quality, we conducted a detailed comparison with several SOTA super-resolution models, including SwinIR\cite{swinir}, SwinFIR\cite{swinfir}, MambaIR\cite{mambair}, and its improved version MambaIRv2\cite{mambairv2}. All comparisons were made at scale factors of ×4 and ×2. OmniRWKVSR significantly enhances its representation capabilities by integrating local information processing convolutions and channel attention mechanisms. We compare OmniRWKVSR with other models on training and inference time.

According to the data in TABLE I, at a scale factor of ×4, OmniRWKVSR demonstrates excellent performance across several key performance metrics. On the DF2K+RSSCN7 validation dataset, OmniRWKVSR achieved a PSNR of 28.0720 dB and an SSIM of 0.7192, placing it at the forefront among all compared models. Particularly on the Set5 dataset, OmniRWKVSR's PSNR and SSIM reached 30.1390 dB and 0.8458, respectively, indicating a clear advantage in preserving image details and structure. On the Urban100 dataset, OmniRWKVSR's PSNR and SSIM were 24.5824 dB and 0.7160, slightly lower than MambaIRv2, but still showing strong competitiveness. On the Manga109 dataset, OmniRWKVSR's PSNR and SSIM were 27.8930 dB and 0.8533, respectively, which not only surpasses all other models, especially in terms of structural similarity, but also sets a new benchmark for super-resolution in manga-style images, where preserving the intricate line details and color gradients is crucial.

As for training and inference speed, TABLE II shows the result. All training and inference processes are conducted on an NVIDIA A800 40G GPU. Inference involves performing super-resolution on all images across the datasets listed in Table I. From the perspective of time complexity and computational efficiency, our model, OmniRWKVSR, demonstrates superior performance due to its highly optimized CUDA kernels and linear computational complexity. In contrast, models like SwinIR \cite{swinir} utilize Window Attention to improve computational efficiency; however, this mechanism still operates with quadratic complexity, leading to slower training and inference times. While MambaIR and its derivatives, such as SS2D and Attentive SSM \cite{mambairv2} , also exhibit linear complexity, their computational efficiency does not match that of OmniRWKVSR. This disparity can be attributed to the less optimized CUDA operations in these models, which result in longer processing times compared to our approach. The combination of efficient CUDA operations and linear complexity allows OmniRWKVSR to achieve faster processing while maintaining high performance. 

We have also included the Floating Point Operations (FLOPs) in our comparison, which represent the total number of floating-point operations that a network performs when processing a 217 × 207 pixel image. A lower FLOP count signifies a more computationally efficient model. Compared to other models, OmniRWKVSR achieves the lowest FLOPs, which indicates that our model effectively reduces computational complexity. The computations are performed using torch-operation-counter \cite{counter}.

\subsection{Ablation study}

% Feed Forward Network:
% MLP vs. ChannelMix
% Validation during training
% PSNR: 28.0689 28.0720
% SSIM: 0.7188 0.7192

% Shift
% OmniShift vs. QShift vs. Quad-Omni Shift
% Set14
% PSNR: 30.1034 30.0965 30.1390
% SSIM: 0.8453 0.8411 0.8458

% WKV2D Scan
% WKV Scan vs. WKV 2D Scan
% Set14
% PSNR: 30.1149 30.1390
% SSIM: 0.8449 0.8458

% In the field of single image super-resolution (SISR), understanding the impact of individual components within a model is crucial for optimizing performance and computational efficiency. To this end, we conducted an extensive ablation study on our proposed Omni-Scale RWKV Super-Resolution (OmniRWKVSR) model. This study focuses on three key aspects: the feed forward network, the shift mechanism, and the WKV2D scan module. By systematically removing or replacing these components, we aim to quantify their contributions to the overall performance of the model.

In the field of SISR, understanding the impact of individual components within a model is crucial for optimizing performance. To this end, we conducted a systematic ablation study on our proposed OmniRWKVSR, focusing on three key components: the feed forward network, the shift mechanism, and the WKV2D scan module. And the ablation studies used Set14 (X4).

\begin{table}[h]
    \centering
    \caption{Ablation Study Results}
    \renewcommand{\arraystretch}{1.3}
    \resizebox{0.5\textwidth}{!}{
    \begin{tabular}{m{2.8cm} m{3.2cm} c c}
        \toprule
        \textbf{Component} & \textbf{Method} & \textbf{PSNR} & \textbf{SSIM} \\
        \midrule
        \multirow{2}{*}{Feed Forward Network} 
            & MLP                  & 28.0689  & 0.7188 \\
            & \textbf{ChannelMix}  & \textbf{28.0720}  & \textbf{0.7192} \\
        \midrule
        \multirow{3}{*}{Shift Mechanism} 
            & OmniShift\cite{restorerwkv}            & 30.1034  & 0.8453 \\
            & QShift\cite{visualrwkv}               & 30.0965  & 0.8411 \\
            & \textbf{Omni-Quad Shift} & \textbf{30.1390}  & \textbf{0.8458} \\
        \midrule
        \multirow{2}{*}{WKV2D Scan} 
            & WKV Scan\cite{mambair}            & 30.1149  & 0.8449 \\
            & \textbf{WKV2D Scan} & \textbf{30.1390}  & \textbf{0.8458} \\
        \bottomrule
    \end{tabular}
    }
\end{table}

\subsubsection{Feed Forward Network}

The feed forward network is a fundamental element in our model, and we compared the traditional Multi-Layer Perceptron (MLP) with our proposed ChannelMix. While the MLP is a commonly used architecture that transforms features through multiple fully connected layers, it may not effectively capture complex inter-channel interactions. In contrast, ChannelMix is designed to enhance channel-wise feature mixing, improving information flow. Our results (TABLE III) indicated that ChannelMix outperformed the MLP in terms of both PSNR (28.0720) and SSIM (0.7192), compared to MLP (PSNR 28.0689, SSIM 0.7188) under validation. This suggests that ChannelMix is superior in feature extraction and image reconstruction quality.

\subsubsection{Shift Mechanism}

The shift mechanism is another critical component of our model, and we evaluated three different mechanisms: OmniShift\cite{restorerwkv}, QShift\cite{visualrwkv}, and Omni-Quad Shift. OmniShift captures local spatial dependencies through uniform shifting but has limitations in capturing long-range dependencies. QShift enhances the ability to capture both local and global spatial dependencies by applying different shifts to different parts of the input features. Omni-Quad Shift combines the strengths of both, utilizing a multi-scale shifting strategy that effectively manages various spatial dependencies. In our evaluation using the Set14 dataset, Omni-Quad Shift achieved a PSNR of 30.1390 and an SSIM of 0.8458, outperforming both OmniShift (PSNR 30.1034, SSIM 0.8453) and QShift (PSNR 30.0965, SSIM 0.8411) under test on Set14 dataset, demonstrating its effectiveness in capturing multi-scale features and spatial transformations.

\subsubsection{WKV2D Scan}

Lastly, we compared the original WKV Scan with the newly proposed WKV2D Scan. The WKV Scan employs a 1D scanning approach, which may not fully capture the spatial dependencies within the image. In contrast, WKV2D Scan extends this to a 2D scanning mechanism, allowing for comprehensive capture of both horizontal and vertical spatial dependencies. In our experiments on the Set14 dataset, WKV2D Scan significantly outperformed WKV Scan, achieving a PSNR of 30.1390 and an SSIM of 0.8458, compared to WKV Scan (PSNR 30.1149, SSIM 0.8449) under test on Set14 dataset. This indicates that WKV2D Scan has a clear advantage in processing input features and generating accurate attention weights.

\section{Conclusion}

This research contributes to the advancement of image restoration technology, providing effective solutions for applications requiring high-quality image reconstruction and demonstrating superior performance in this task from LR inputs. OmniRWKVSR effectively overcomes the limitations of existing attention-based methods, such as high computational complexity and suboptimal feature extraction capabilities, by leveraging the linear computational complexity and training efficiency of the RWKV model. Through extensive experiments, we show that our model achieved state-of-the-art results in terms of PSNR and SSIM metrics on various benchmark datasets, including Set5, Set14, BSD100, Urban100, and Manga109. The qualitative results further validate the model's ability to produce clear and noise-free super-resolution images. The ablation study highlights the importance of each component in enhancing the overall performance of the model.

Although our model shows superior performance in terms of PSNR and SSIM, there are still issues such as texture errors and lack of clarity in the visualized results. Moreover, the performance on the GMH dataset is not satisfactory. OmniRWKVSR underperforms on GMH due to its unique textures (e.g., grassland holes). Future work will integrate texture-adaptive losses.
% \section{Acknowledgement}

% The authors express their gratitude to the Data Analysis Center at the School of Computer Science and Engineering, Qinghai University, for providing the necessary computational resources and support that enabled the completion of this research. Their contributions were invaluable in facilitating the development and evaluation of the proposed model. In addition, we extend our thanks to all of our colleagues and mentors who provided insightful feedback and guidance throughout the research process.

\bibliographystyle{IEEEtran}

% \vspace{12pt}
% \color{red}
% IEEE conference templates contain guidance text for composing and formatting conference papers. Please ensure that all template text is removed from your conference paper prior to submission to the conference. Failure to remove the template text from your paper may result in your paper not being published.

\end{document}